\documentclass{svproc}
\usepackage{makeidx}  % allows for indexgeneration
\usepackage{cite}
\usepackage{graphicx}
\usepackage{todonotes}
\usepackage{tabularx}
\graphicspath{ {./figures/} }
\usepackage{etoolbox} % for "\patchcmd" macro
\makeatletter
\patchcmd{\ps@headings}{\rlap{\thepage}}{}{}{}
\patchcmd{\ps@headings}{\llap{\thepage}}{}{}{}
\makeatother
\begin{document}
%
%\frontmatter          % for the preliminaries
%
%\pagestyle{headings}  % switches on printing of running heads
%\addtocmark{Hamiltonian Mechanics} % additional mark in the TOC

%\tableofcontents
%
\mainmatter              % start of the contributions
\title{Gradient based Grasp Pose Optimization on a NeRF that Approximates Grasp Success}
\titlerunning{Grasp Pose Optimization with NeRF}
\author{Gergely S\'{o}ti\inst{1, 2} \and
Björn Hein\inst{1, 2}  \and 
Christian Wurll\inst{1}%
}
\authorrunning{Gergely S\'{o}ti et al.} % abbreviated author list (for running head)
\institute{
Hochschule Karlsruhe -- University of Applied Sciences, 76133 Karlsruhe, Germany\\
\and
Karlsruhe Institute of Technology, 76131 Karlsruhe, Germany \\
\email{gergely.soti@h-ka.de}\thanks{This research is being conducted as part of the KI5GRob project funded by the German Federal Ministry of Education and Research (BMBF) under project number 13FH579KX9.}}

\maketitle              % typeset the title of the contribution

\begin{abstract}
Current robotic grasping methods often rely on estimating the pose of the target object, explicitly predicting grasp poses, or implicitly estimating grasp success probabilities. In this work, we propose a novel approach that directly maps gripper poses to their corresponding grasp success values, without considering objectness. Specifically, we leverage a Neural Radiance Field (NeRF) architecture to learn a scene representation and use it to train a grasp success estimator that maps each pose in the robot's task space to a grasp success value. We employ this learned estimator to tune its inputs, i.e., grasp poses, by gradient-based optimization to obtain successful grasp poses. Contrary to other NeRF-based methods which enhance existing grasp pose estimation approaches by relying on NeRF's rendering capabilities or directly estimate grasp poses in a discretized space using NeRF's scene representation capabilities, our approach uniquely sidesteps both the need for rendering and the limitation of discretization. We demonstrate the effectiveness of our approach on four simulated 3DoF (Degree of Freedom) robotic grasping tasks and show that it can generalize to novel objects. Our best model achieves an average translation error of 3mm from valid grasp poses. This work opens the door for future research to apply our approach to higher DoF grasps and real-world scenarios.


\end{abstract}
\section{Introduction}

Research in robotic grasping has explored various approaches such as analytic and data-driven, model-based and model-free, supervised, self supervised and reinforcement learning methods. These methods can be based on different types of sensor data, such as RGB or depth images, and can be designed for different types of grippers \cite{kleeberger2020survey}.

Most of these methods are based on object pose estimation, directly estimate a grasp pose or implicitly map grasp poses to their probability of success.
However, if we observe ourselves while grasping an object, we might notice, that we intuitively adjust our hand position to increase the chances of a successful grasp and to achieve a good grasp position ultimately. This suggests that the process of grasping can be modeled as an optimization problem that optimizes the pose of our hands to maximize the probability of a successful grasp.

In this work, we introduce a novel approach to robotic grasping. Leveraging VisionNeRF \cite{lin2023vision}, a learned neural network model capable of capturing a 3D scene representation, we create a model that estimates the success of a grasp given a candidate pose. Unlike other -- including NeRF-based -- grasping methods which directly estimate grasp poses, our approach stands out by formulating grasp pose estimation as a continuous optimization problem. The goal is to maximize the likelihood of successful grasping through gradient-based optimization.  We show the efficiency of our proposed approach on four simulated 3DoF robotic grasping tasks. We summarize our contributions as follows:
\begin{itemize}
    \item We propose a method to explicitly map grasp candidates to their corresponding grasp success value.
    \item We show the efficacy of applying transfer learning to a trained VisionNeRF to obtain this explicit mapping.
    \item We propose a novel approach to find valid grasp poses by applying gradient based optimization on the learned grasp success estimator. 
\end{itemize}

\section{Related Work}

\subsection{Data-driven Robotic Grasping}
In recent years, data-driven methods have become the state of the art in the context of robotic object handling. Keypoint detection or dense descriptor-based methods are effective at learning successful grasp poses and can even generalize to object categories, but they often require a large amount of object-specific labeled data to achieve good performance \cite{florence2018dense, manuelli2019kpam, kulkarni2019unsupervised, nagabandi2020deep, liu2020keypose}. End-to-end learning models that directly learn to map the robot's raw sensor input to a desired output offer a promising alternative in unstructured environments \cite{wu2019learning, zakka2020form2fit, hundt2020good, berscheid2020self, devin2020self, khansari2020action, song2020grasping, zeng2020transporter}. Most of these models directly propose suitable grasp candidates, or estimate the success probability of grasp poses and rank them. These latter models implicitly map grasp poses to success probability, limiting their ability to optimize grasp poses to iterative methods \cite{soti2023train} that sample, evaluate, and re-sample grasp candidates to find a better solutions. In contrast, our proposed method explicitly maps grasp poses to grasp success using a neural network, making it differentiable and enabling gradient-based optimization to refine the grasp pose.

\subsection{NeRFs and NeRF-based Robotic Grasping}
Recently, differentiable scene representations, such as Neural Radiance Fields \cite{mildenhall2020nerf}, have been increasingly used in the field of robotics also for grasping among other applications.
A NeRF maps a 5-degree-of-freedom (5DoF) pose to an RGB color vector and a so-called density. Color and density are then combined along camera rays via volumetric rendering in order to render novel views for scenes. Various extensions of the NeRFs have been developed for different applications, such as NeuS \cite{wang2021neus} for surface reconstruction or NeRF-W \cite{martin2021nerf} on unconstrained photo collections of famous landmarks. Plenoctrees \cite{yu2021plenoctrees} have been proposed for fast rendering with NeRFs. PixelNeRF \cite{yu2021pixelnerf} and VisionNeRF \cite{lin2023vision} overcome the need for training a NeRF for each scene, by generalizing over multiple scenes given sparse observations.

Inverse Neural Radiance Fields \cite{yen2021inerf} perform camera pose estimation by inverting a trained NeRF. Starting from an initial camera pose estimate, it uses gradient based optimization to minimize the residual between pixels rendered from an already-trained NeRF and pixels in an observed image. To estimate the 6DoF camera pose, iNeRF casts rays from the camera's perspective and samples points along them, to finally apply volumetric rendering to get the pixel values and thus the residual. This requires querying the NeRF with different 5DoF poses multiple times. In our method, we use a similar approach, but since we are only interested in estimating 3DoF poses (5DoF with a fixed direction), we can simply use the NeRF's output at 5DoF poses as an objective.

There are several successful methods that utilize variants NeRFs for robotic grasping. Dex-NeRF \cite{ichnowski2021dex} uses a NeRF-based model to render high-quality depth images of a scene, which are then fed to DexNet \cite{mahler2017dex} to compute robust grasp poses. Evo-NeRF \cite{kerrevo}, is similar method, but instead of focusing on improving the depth rendering, the grasp planner network is trained to perform well on the NeRF-rendered depth maps and utilizes a different NeRF implementation to significantly improve training times. GraspNeRF \cite{dai2022graspnerf} utilizes a multiview NeRF-based approach to estimate a truncated-signed-distance-field in voxels to predict successful grasps. An other approach \cite{yen2022nerf} uses a NeRF to learn dense object descriptors from visual observations, which are then used to track keypoints on objects and calculate grasp poses.

\subsection{Our Contribution}

While these methods demonstrate the effectiveness of utilizing NeRFs in robotic grasping, they typically enhance existing grasp pose estimation techniques with NeRFs' rendering capabilities or directly estimate grasp poses in a discretized space using NeRFs. However, these methods do not fully exploit the potential of NeRFs for continuous optimization of the grasp pose.

Our method uniquely employs NeRFs to explicitly represent the mapping of grasp poses to grasp success probability. This approach enables a gradient-based optimization method to find optimal grasp poses, providing more fine-grained control over the optimization process. Furthermore, our explicit mapping of grasp poses to grasp success offers a natural representation of the problem, where the gradient directly depicts rigid transformations leading to more successful poses. We believe our method addresses the gaps in the current state of the art and introduces a fresh perspective to the field of robotic grasping.

\section{Grasp Success Approximation and Optimization}

Given an RGB observation of a tabletop scene, the goal of the proposed method is to detect 5-DoF grasps (e.g. with a suction cup) consisting of a position and a direction vector. We assume the camera intrinsics and extrinsics are known for the image. We formulate the 5-DoF grasp detection as an optimization problem, that maximizes grasp success probability over gripper poses. We approximate the function that maps 5-DoF grasp poses $g = (x, d) \in \mathbf{G}$, with $x$ position and $d$ direction, to their probability of success by the neural network $\mathbf{\Theta}$. Since neural networks are differentiable, we can solve the problem

\begin{equation}
    \max_{g \in \mathbf{G}} \mathbf{\Theta}(g, o)
\end{equation}
by gradient based optimization methods, where $o = (c, K, RT)$ is an observation containing a camera image with known intrinsics and extrinsics.

In this section we first describe the architecture of $\mathbf{\Theta}$ and how we train it, then we describe the gradient based optimization.
Note, that although this formulation is valid for 5DoF grasps, we constrain ourselves to 3DoF graps (position only with fixed direction) in the evaluation.

\subsection{Grasp Success Approximation}
NeRFs excel in novel view synthesis and are increasingly being applied in various other tasks that require some sort of scene representation. By using volumetric rendering to compute the loss during training, NeRFs are forced to learn how to consistently represent 3D scenes. In this paper, we demonstrate the potential of this representation for grasp success estimation.
\subsubsection{Architecture}
In our approach we use a VisionNeRF \cite{lin2023vision}, a generalized implementation of NeRFs capable of representing multiple scenes by conditioning on observed inputs. To achieve this, a Vision Transformer (ViT) \cite{dosovitskiy2020image} and a Convolutional Neural Network (CNN) are combined to extract global and local features from the input observation, the source image, which are then used to inform the color and density estimator. We denote this combination as $\mathbf{\Omega}$. While standard NeRFs map a 5DoF pose $(x, d)$(corresponding to a 3D point in the scene and the camera's perspective) to an RGB color vector and density, VisionNeRFs require an additional input: a single camera image $c$ of the scene with known intrinsics $K$ and extrinsics $RT$.

\begin{figure}[ht]
    \vspace{-0.3cm}
    \centering
    \includegraphics[width=0.7\textwidth]{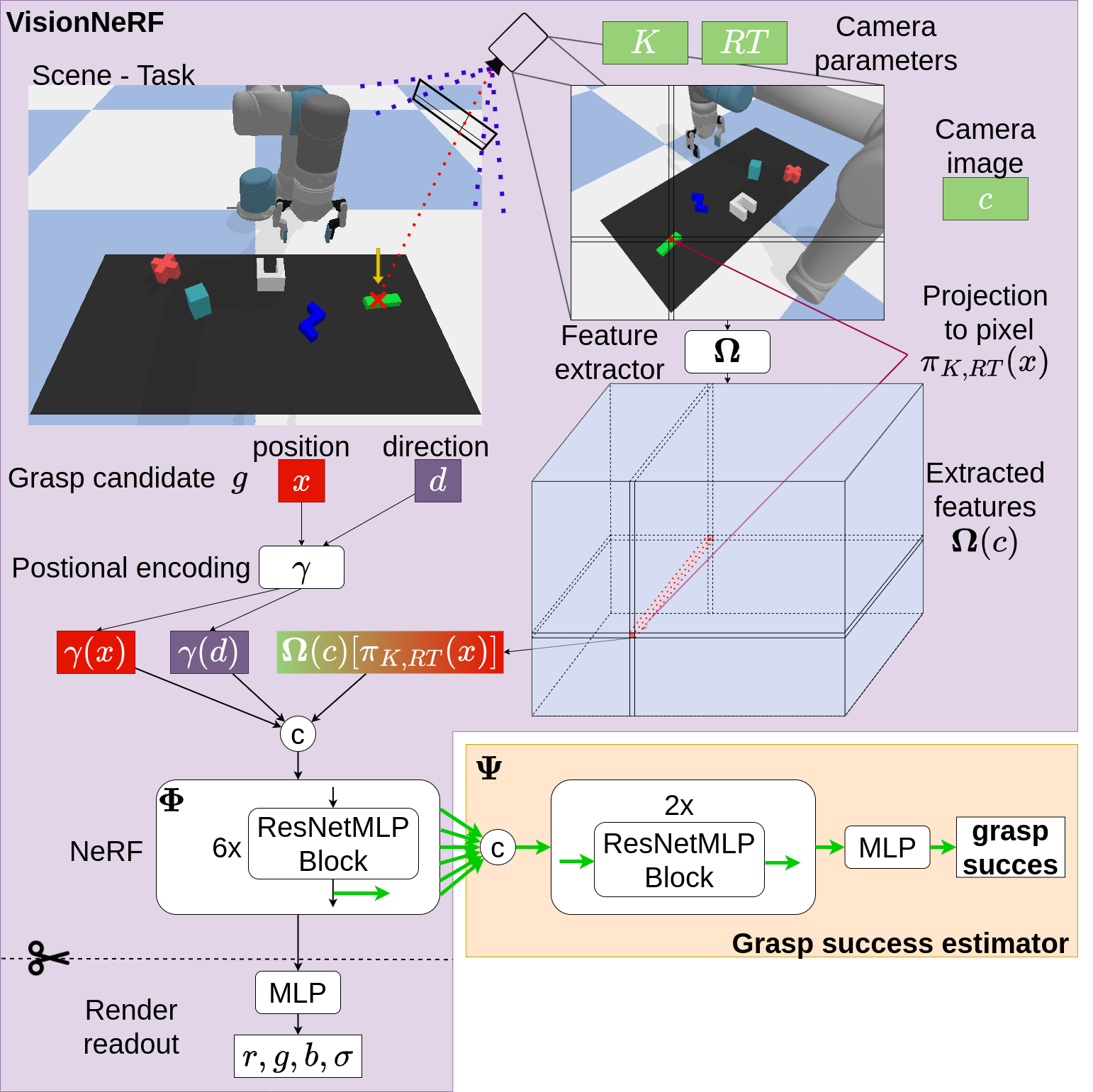}
    \caption{The structure of our proposed architecture: a VisionNeRF that estimates color and density for volumetric rendering and a grasp success estimator. Both process 5DoF poses $(x, d)$ and are informed by the extracted features from the camera image $c$ that correspond to $x$}
    \vspace{-0.3cm}
    \label{fig:grasp_success_architecture}
\end{figure}

The NeRF architecture consists of a sequence of ResNetMLP blocks (see \cite{lin2023vision}, for more details) denoted as $\mathbf{\Phi}$, and a final fully connected layer that outputs the color and density of the 3D point $x$ given an observation direction $d$. To inform this color and density estimator, the feature vector from $\mathbf{\Omega}(c)$ at the projected position of the 3D point onto the camera image $\pi_{K,RT}(x)$ is concatenated with encoded $x$ and $d$ vectors. We use a positional encoding typical for NeRFs:
\begin{equation}
    \gamma(p) = (sin(2^0\pi p), cos(2^0\pi p), ..., sin(2^{M-1}\pi p), cos(2^{M-1}\pi p))
\end{equation}
with $M$ the number of frequency phases. Note, VisionNeRF only applies position encoding to the position vector and not the direction vector, but we also use it on the direction vector, just like the original NeRF implementation.
This concatenated vector of $\gamma(x)$, $\gamma(d)$ and $\mathbf{\Omega}(c)[\pi_{K,RT}(x)]$ is then fed into $\mathbf{\Phi}$ and passed to the final fully connected layer. The output colors and densities of multiple points along a camera ray are then integrated using volumetric rendering to obtain pixel values, thus rendering the target image, as shown in Fig. \ref{fig:grasp_success_architecture}.

To leverage the learned representation, we propose an extension to the VisionNeRF architecture. One potential issue is that the output of $\mathbf{\Phi}$, which is primarily trained to approximate color and density, can be biased towards these features. To address this, we introduce skip connections after each ResNetMLP block in $\mathbf{\Phi}$. By concatenating these skip connections with the output of $\mathbf{\Phi}$, we create the input for our grasp success estimator module, denoted as $\mathbf{\Psi}$. $\mathbf{\Psi}$ consists of two ResNetMLP blocks and a final fully connected layer, which outputs a grasp success score. Fig. \ref{fig:grasp_success_architecture} shows the architecture of our model, but for visualization purposes we only depict the position and the direction of our grasp candidate $g = (x, d)$ in the image. In reality we propagate four 5DoF poses through the network simultaneously, all along $d$ and centered around $x$ with 2.5mm spacing. We sum up the output of the grasp success estimator for these poses to obtain the final grasp success of the $g$.

With these we can define the objective function of the optimization problem:
\begin{equation}
    \mathbf{\Theta}(g, o) = \mathbf{\Psi}(\mathbf{\Phi}(\gamma(x), \gamma(d), \mathbf{\Omega}(c)[\pi_{K,RT}(x)]))
\end{equation}
with grasp candidate $g=(x,d)$ and observation $o = (c, K, RT)$.

\subsubsection{Training}
To get the model to learn to represent the scene, we initially train $\mathbf{\Omega}$ and $\mathbf{\Phi}$ for novel view synthesis via volumetric rendering. The ViT in $\mathbf{\Omega}$ is initialized with pretrained weigths from \cite{rw2019timm}. For training we use the Adam optimizer with a warmup learning rate schedule. The learning rate is increased from 0 to 1e-4 in 10000 steps for $\mathbf{\Omega}$ and similarly for $\mathbf{\Phi}$, the learning rate is increased from 0 to 1e-5 in 10000 steps.

After training we apply transfer learning to the VisionNeRF by freezing the weights of $\mathbf{\Omega}$ and $\mathbf{\Phi}$ and training only $\mathbf{\Psi}$ to obtain the complete grasp success estimator $\mathbf{\Theta}$. Categorical cross-entropy loss is used with one successful grasp pose $g$ for an observation $o$ as a positive example labeled as 1 and multiple randomly sampled poses as negative examples from the workspace labeled as 0. To obtain a valid grasp pose, we sample a position $x$ on the top surface of the (prismatic) object - the optimal site for a suction gripper. We set the direction $d$ perpendicular to this surface. We use the Adam optimizer with learning rate 1e-4 and sample 2047 negative samples.

As baseline, we use the same architecture, but instead of pretraining $\mathbf{\Omega}$ and $\mathbf{\Phi}$, we only load the ViT pretrained weights. We then train $\mathbf{\Omega}$, $\mathbf{\Phi}$ and $\mathbf{\Psi}$ jointly, with the same configurations as described above.

\subsection{Gradient based Optimization}
\label{subsec:g-b-o}
To solve the optimization problem, we used a gradient base optimization method. We apply the Adam optimizer with a decaying learning rate starting at 0.05 with decay rate 0.8 after each step to minimize the objective function $-\mathbf{\Theta}(g)$ over $g \in \mathbf{G}$ grasp poses, thus maximizing the estimated grasp success. We initialize the optimization process with $2^{13}$ random poses as grasp candidates, and the optimization is allowed to run for a maximum of 16 iterations.

Since we constrain ourselves to 3DoF poses only, we fix the direction $d$ and only optimize $x$. This constraint is also applied while training the grasp optimizer by only sampling negative examples with the same direction. 

In the context of grasping, the gradient used for optimization corresponds to rigid transformations of the gripper that lead to more successful grasp poses.

\section{Experiments}

\begin{figure}[ht]
    \vspace{-0.3cm}
    \centering
    \includegraphics[width=0.8\textwidth]{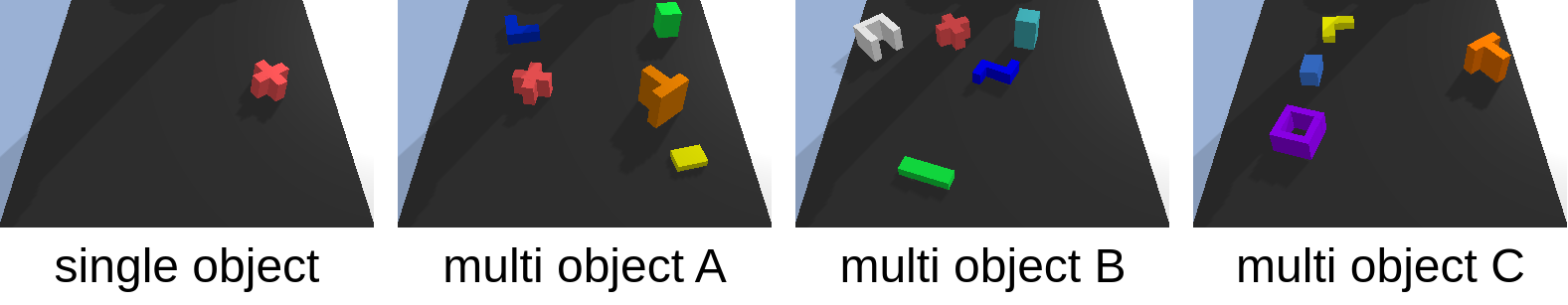}
    \caption{The different object sets used during training and evaluation.}
    \vspace{-0.3cm}
    \label{fig:object_sets}
\end{figure}

We use a simulated tabletop environment to evaluate the performance of the proposed approach on 3DoF robotic grasping tasks. There are three fixed-pose cameras in the environment that provide camera images as observations. To measure the accuracy of the grasping predictions, we computed the translation error, which represents the distance between a predicted grasp position and the nearest valid grasp position. Our approach enables the simultaneous optimization of multiple grasp candidates by maximizing their predicted success rate $\mathbf{\Theta}(g, o)$. We evaluated its performance using two different metrics: 
\begin{itemize}
    \item \textbf{best-success}: the translation error of the grasp with the highest predicted success rate
    \item \textbf{lowest-from-5}:the lowest translation error among the five grasp candidates with the highest predicted success rates, which can be roughly understood as if a grasp fails, we can try the next best candidate
\end{itemize}
For a task, we spawn objects from one of the following sets of objects (see Figure \ref{fig:object_sets}):
\begin{itemize}
    \item \textbf{single object}: red cross (0.05)% (see Fig. \ref{fig:grasp_success_estimation})
    \item \textbf{multi object A}: red cross (0.05), green square (0.07), yellow rectangle (0.015), dark blue L-shape (0.03), orange T-shape (0.09)% (see Fig. \ref{fig:nerf_rendering})
    \item \textbf{multi object B}: red cross (0.05), turquoise square (0.08), green long rectangle (0.02), white U-shape (0.06), dark blue double-L-shape (0.03)% (see Fig. \ref{fig:grasp_success_architecture})
    \item \textbf{multi object C}: blue rectangle (0.04), yellow L-shape (0.02), orange T-shape (0.07), purple block-ring (0.05)
\end{itemize}

All objects are prismatic and are characterized by their bases, heights (in meters) and colors. The set multi object A contains objects similar to some of the other multi object sets, while multi object B and multi object C contain mainly different objects.

We define two tasks in which objects need to be grasped:
\begin{itemize}
\item \textbf{single object grasp}: the object of the single object set is spawned in a random position in the workspace
\item \textbf{multi object grasp}: five objects are sampled from a multi object set and spawned in random non-overlapping positions in the workspace; the objects need to be removed successively one by one resulting in 5 different scenes for a complete episode
\end{itemize}

In our experiment we use three different backbones:
\begin{itemize}
    \item \textbf{no-NeRF}: the baseline without pretraining the VisionNeRF
    \item \textbf{single-NeRF}: a VisionNeRF trained on 100 scenes of the single object grasp task, corresponding to 100 complete episodes of the task
    \item \textbf{multi-NeRF}: a VisionNeRF trained on 500 scenes of the multi object grasp task with objects from the multi object A set, corresponding to 100 complete episodes of the task
\end{itemize}
Both single-NeRF and multi-NeRF are trained for 8000 epochs with batch-size 1. Source and target camera images are sampled from the three fixed-pose camera observations. In contrast, NeRFs are generally trained using many different views which is not realistic in real world setups.

With all three backbones, we train two grasp success estimator modules:
\begin{itemize}
    \item \textbf{single-grasp}: trained on 100 scenes of the single object grasp task, corresponding to 100 complete demonstrations of the task
    \item \textbf{multi-grasp}: trained on 100 scenes of the multi object grasp task with objects from the multi object B set, corresponding to 20 complete demonstrations of the task,
\end{itemize}
resulting in six models overall. All grasp estimator modules are trained for 250 epochs. The models are evaluated on four tasks:
\begin{itemize}
    \item \textbf{single-object-task}: 50 scenes of the single object grasp task using the single object object set, corresponding to 50 complete episodes
    \item \textbf{multi-object-A-task}: 50 scenes of the multi object task using objects from the multi object A object set corresponding to 10 complete episodes; note, that these objects were also used for training the multi-NeRF module
    \item \textbf{multi-object-B-task}: 50 scenes of the multi object task using objects from the multi object B object set corresponding to 10 complete episodes; note, that these objects were also used for training the multi-grasp module
    \item \textbf{multi-object-C-task}: 50 scenes of the multi object task using objects from the multi object C object set corresponding to 10 complete episodes
\end{itemize}
For each task we obtain three observations $o_1$, $o_2$ and $o_3$, one for each camera. For $\mathbf{\Theta}(g, o)$ however, we only need one observation, thus we define two optimization objectives:
\begin{itemize}
    \item \textbf{1-view}: $\mathbf{\Theta}(g, o_1)$
    \item \textbf{3-views}: $\sum_{i \in [1, 2, 3]}{\mathbf{\Theta}(g, o_i)}$
\end{itemize}
We record the best five grasp candidates with the highest estimated grasp success score after 8, 12 and 16 optimization steps for evaluation.

\section{Results and Discussion}
\subsection{Qualitative Analysis of Architecture Modules}
\subsubsection{VisionNeRF} 
\begin{figure}[ht]
    \vspace{-0.3cm}
    \centering
    \includegraphics[width=\textwidth]{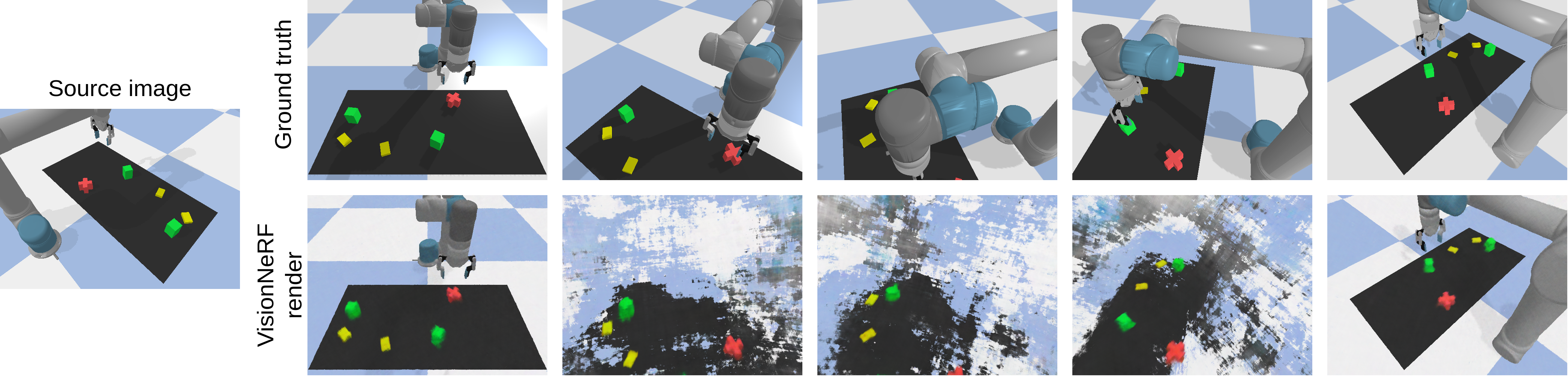}
    \caption{VisionNeRF rendering of new perspectives given a source image with known perspective. The left- and right-most renderings belong to perspectives that were used at training and produce better quality images. For the other perspectives, the static objects in the scene (ground and robot) seems to fall apart, but the objects are rendered consistently even if they were ocluded in the ground truth images.}
    \vspace{-0.3cm}
    \label{fig:nerf_rendering}
\end{figure}

As described above, we trained our VisionNeRF for novel view synthesis only using three perspectives. This leads to a strong bias towards these perspectives during rendering new perspectives given a camera image from a known perspective. When rendering for perspectives that were used during training, the quality of the image is far superior than for other perspectives, however the representation of the objects in the workspace remains mostly consistent, as shown in Fig. \ref{fig:nerf_rendering}.

\subsubsection{Grasp success estimator and grasp pose optimization} 
\begin{figure}[ht]
    \vspace{-0.3cm}
    \centering
    \includegraphics[width=0.6\textwidth]{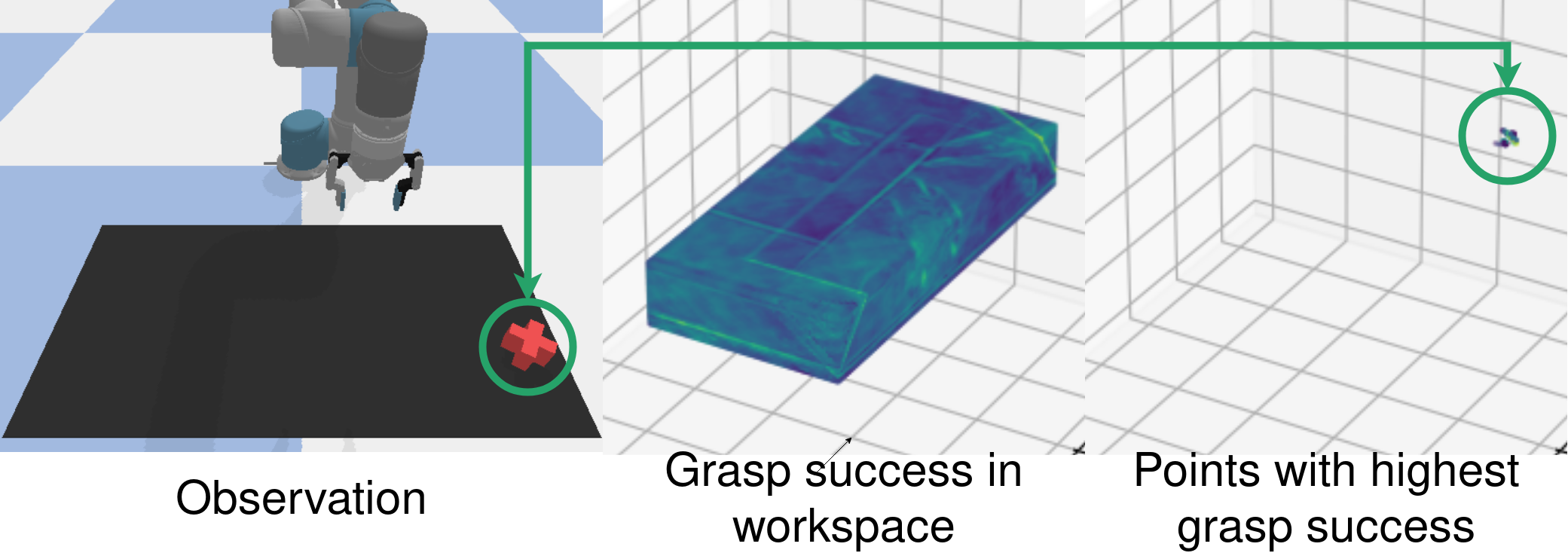}
    \caption{A visualization of the grasp success estimation: the discretized workspace of an instance of the single-grasp-task (left) is mapped to its 3-views optimization objective $\sum_{i \in [1, 2, 3]}{\mathbf{\Theta}(g, o_i)}$ (middle). On the right, only the points with the highest success estimation values are shown, also corresponding to the object's position in the workspace.}
    \label{fig:grasp_success_estimation}
\end{figure}

To ensure that our grasp pose estimation method, which involves gradient-based optimization, is accurate, the learned grasp success estimation function must correctly map 3D (or 5D) space to grasp success. Ideally, the function should assign higher success estimates to points closer to valid grasp positions. Although our learned grasp success estimator has some local maxima that do not correspond to valid grasp poses, the global maxima do, as shown in Fig. \ref{fig:grasp_success_estimation}. 

\begin{figure}[ht]
    \centering
    \includegraphics[width=0.6\textwidth]{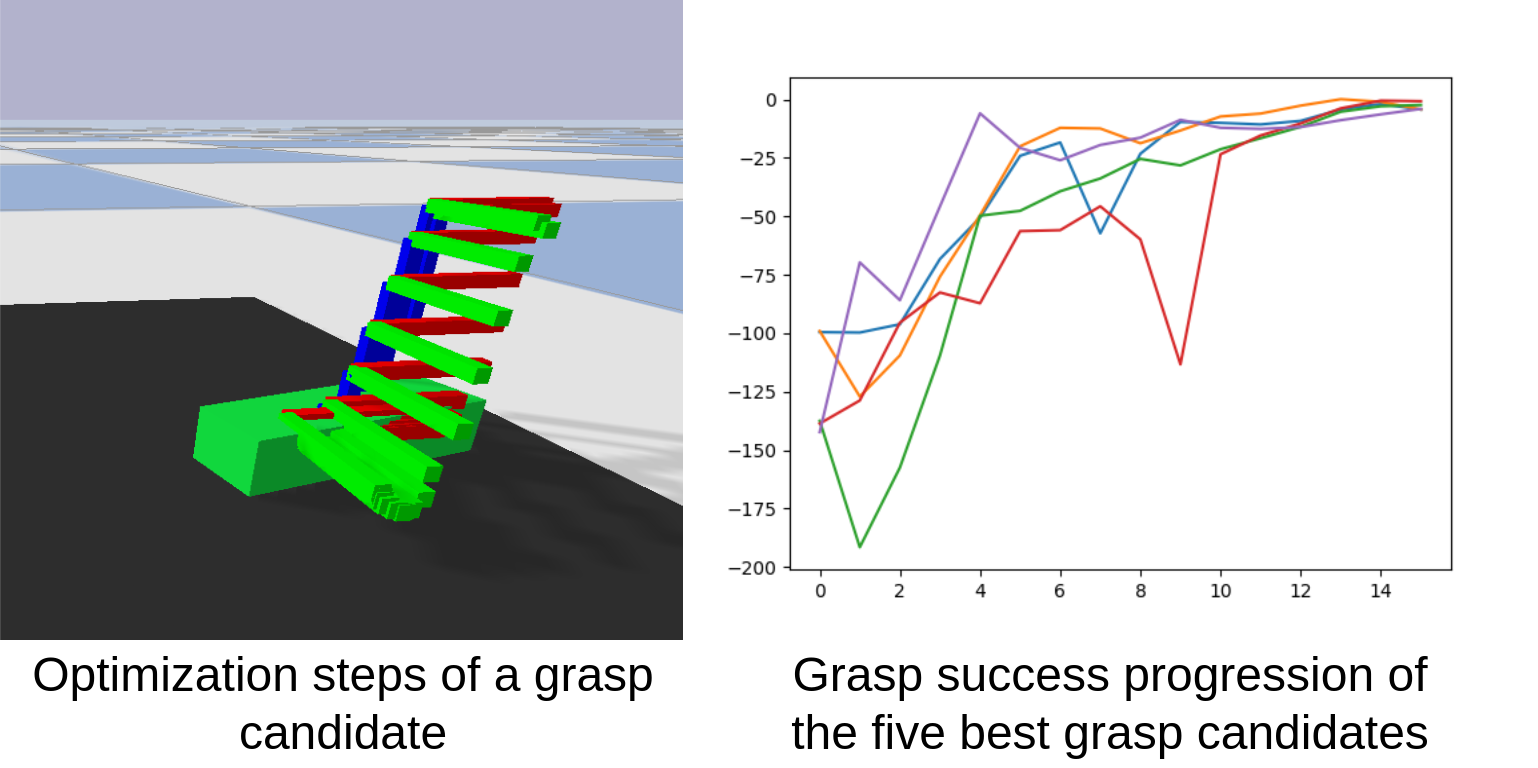}
    \caption{Grasp pose improvement during optimization (left) and the estimated grasp success progression of the five grasp candidates that have the highest estimated grasp success at the end of the optimization (right).} 
    \vspace{-0.3cm}
    \label{fig:grasp_optimization}
\end{figure}

Of course, gradient based optimization methods are prone to get stuck in local limits. We overcome this problem by initializing the optimization method with many initial grasp candidates as described in \ref{subsec:g-b-o} and evaluating the grasp candidates that have the highest grasp success estimation at the end of the optimization. Fig. \ref{fig:grasp_optimization} shows the successively improved poses of a grasp candidate during optimization and the estimated grasp success progression of the grasp candidates with the highest estimated grasp success at the end of the optimization. This also suggests, that the gradient does indeed correspond to a rigid transformation that moves the gripper towards better grasp poses.

\subsection{Robotic Grasping Performance Evaluation}
\begin{figure}[ht]
    \vspace{-0.3cm}
    \centering
    \includegraphics[width=1.0\textwidth]{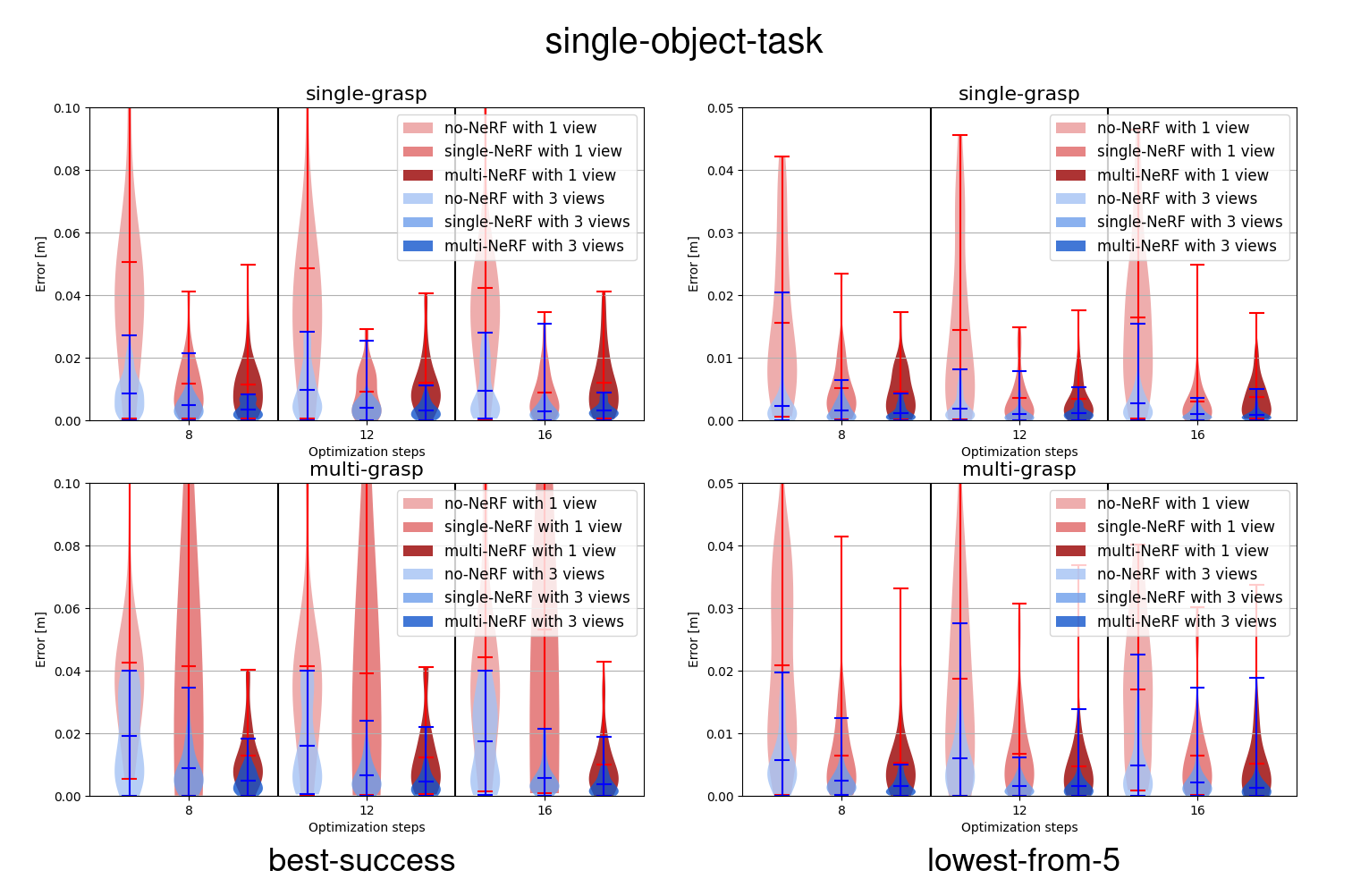}
    \caption{Error distribution of different model configurations on the single-object-task after 8, 12, and 16 optimization steps using 1-view and 3-views as optimization objective for the metrics best-success and lowest-from-5.}
    \vspace{-0.3cm}
    \label{fig:single_object_eval}
\end{figure}

Using 3-views instead of 1-view for the optimization objective reduces the errors of our approach by over an average of 60\% for all backbone and grasp success estimator combinations as shown in Fig. \ref{fig:single_object_eval} using both best-success and lowest-from-5 metrics. Furthermore, for both optimization objectives, the architectures with a pretrained NeRF outperform the models that did not make use of transfer learning, while models with multi-NeRF mostly even outperform models with single-NeRF. A significant exception is observable when models using multi-grasp are evaluated with the best-success metric, where single-NeRF architectures do not outperform models that do not use a pretrained NeRF backbone. This suggests that using a single view in the objective does not depict the reality as reliably as using three views.

In the single-object-task (Fig. \ref{fig:single_object_eval}), architectures that combine a single-NeRF with a single-grasp models can slightly outperform their multi-NeRF counterparts, which is however reasonable, as both single-NeRF and single-grasp models were trained on the same object set, that is used in this task.

\begin{figure}[ht]
    \vspace{-0.3cm}
    \centering
    \includegraphics[width=\textwidth]{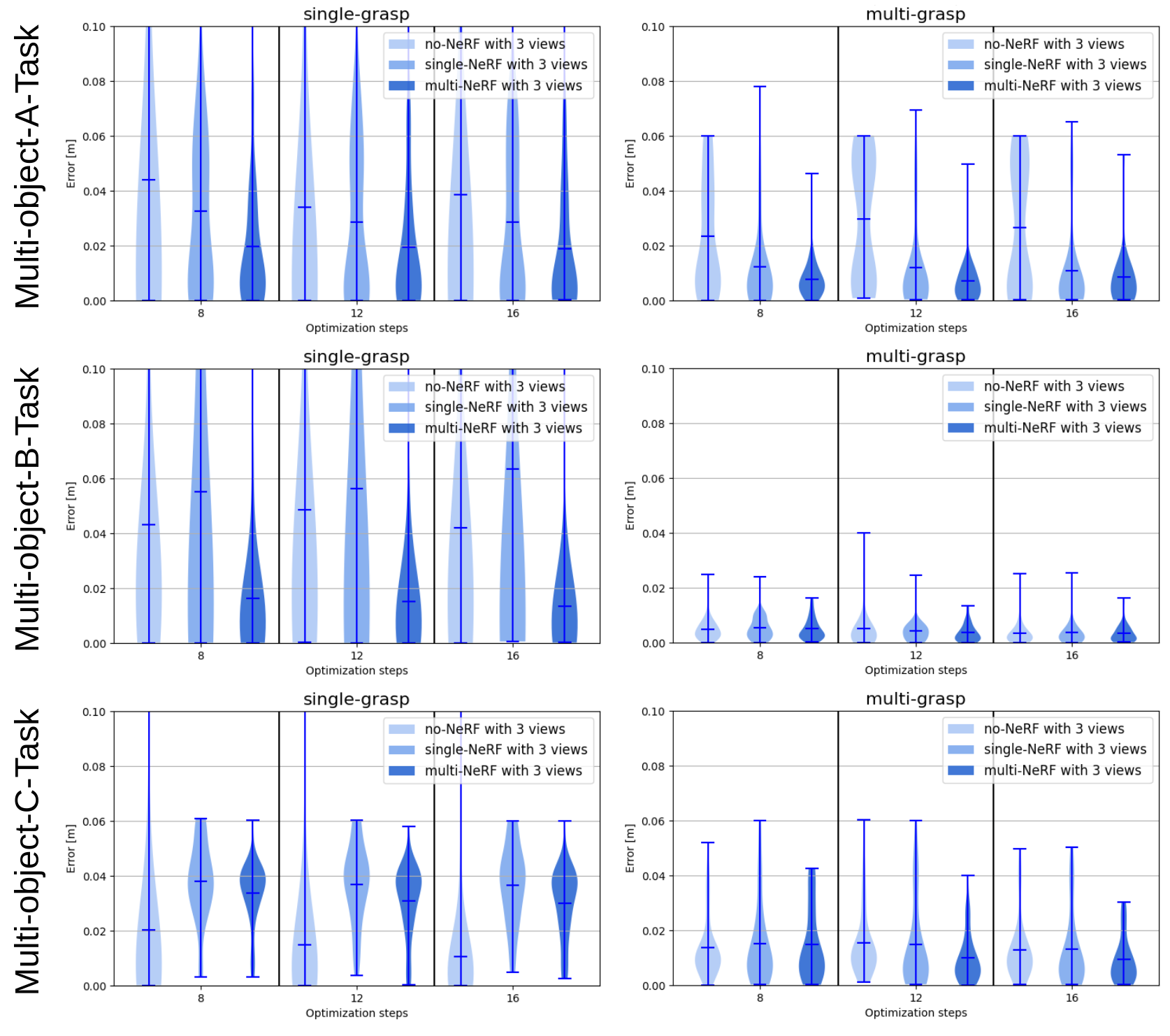}
    \caption{Best-success error distribution of different model configurations on all multi-object-tasks after 8, 12, and 16 optimization steps using 3-views as optimization objective.}
    \vspace{-0.3cm}
    \label{fig:multi_object_eval}
\end{figure}

We can observe similar behaviour if we observe the results of multi-grasp models with different backbones on the different multi-object-tasks (Fig. \ref{fig:multi_object_eval}, right): that models using a NeRF backbone mostly outperform models without a pretrained NeRF backbone and that multi-NeRF outperforms single-NeRF in most cases. Additionally, all models perform best on the multi-object-B-task, which is again reasonable, as the multi-grasp models were trained on the same object set. In case of the multi-object-A-task, the model using multi-NeRF clearly outperforms the other models, which is most likely due to the fact, that the multi-NeRF was also trained on multi object A object set, thus it likely extracts the most descriptive features from scenes with these objects. In the multi-object-C-task the no-NeRF end-to-end model and the single-NeRF model perform similarly and are still outperformed by the multi-NeRF architecture. 

When we examine single-grasp models on the same tasks (Fig. \ref{fig:multi_object_eval}, left), only multi-object-A-task shows the same pattern regarding backbone configuration. For the multi-object-B task, the architecture without a pretrained NeRF outperforms the single-NeRF architecture, while the model with a multi-NeRF backbone still outperforms both. In case of multi-object-C-task however, the end-to-end architecture delivered the best results.

% rows: ['suction-single-test', 'suction-multi-a-test', 'suction-multi-b-test', 'suction-multi-test'] x low, best

On average, our models achieved best performance after 16 optimization steps, but only slightly better than after 12 optimization steps. Table \ref{table:1} summarizes the average errors (in mm) for all models after 16 optimization steps. Considering the best-success metric, for the single-object-task, the single-NeRF with a single-grasp performs best, although it is worth noting, that both multi-NeRF based models have less than 0.7mm larger average error. In case of all multi-object-tasks, the multi-NeRF and the multi-grasp model show better performance, albeit in the multi-object-B-task only slightly better than the other models with a multi-grasp module.

The lowest-from-5 metric models the case when we also consider retrying a failed grasp. As Table \ref{table:1} demonstrates, the results show similar trends, though not as consistent as for the best-success metric. There is one major outlier: in case of the multi-object-C-task the model combining a single-NeRF with a single-grasp model outperforms all other combinations.

% cols: single [no single multi]; multi [no, single, multi]
\setlength{\tabcolsep}{2.pt}
    \vspace{-0.3cm}
\begin{center}
\begin{table}
\caption{\label{table:1} Average errors of all models using the 3-views optimization objective in mm according to the best-success and lowest-from-5 errors. The single-object-task is denoted as so and the different multi-object-tasks are denoted with mo-X with X referring to the obeject set they were defined on.}
\begin{tabular}{l || c| c | c | c | c | c}
\cline{2-7}
\multicolumn{1}{c}{}& \multicolumn{6}{c}{\textbf{best-success}} \\
\cline{2-7}
 \multicolumn{1}{c}{}& \multicolumn{3}{c|}{single-grasp} & \multicolumn{3}{c}{multi-grasp}\\
\cline{2-7}
 \multicolumn{1}{c}{}& no-NeRF & single-NeRF & multi-NeRF & no-NeRF & single-NeRF & multi-NeRF \\
\hline
so & 9.39 &  \textbf{2.94} &  3.17 & 17.50 &  5.68 & 3.61 \\
mo-A & 38.46 & 28.43 & 18.81 & 26.47 & 10.73 & \textbf{8.46} \\
mo-B & 41.98 & 63.30 & 13.34 &  3.43 &  3.59 & \textbf{3.41} \\
mo-C & 10.67 & 36.50 & 29.98 & 12.86 & 13.09 & \textbf{9.33} \\
\hline
\hline
\multicolumn{1}{c}{}& \multicolumn{6}{c}{\textbf{lowest-from-5}} \\
\cline{2-7}
 \multicolumn{1}{c}{}& \multicolumn{3}{c|}{single-grasp} & \multicolumn{3}{c}{multi-grasp}\\
\cline{2-7}
 \multicolumn{1}{c}{}& no-NeRF & single-NeRF & multi-NeRF & no-NeRF & single-NeRF & multi-NeRF \\
\hline
so & 2.70 &  1.05 &  \textbf{0.91} &  4.87 &  2.16 & 1.31 \\
mo-A & 22.22 & 22.87 & 13.67 & 13.63 &  5.75 & \textbf{3.40} \\
mo-B & 28.45 & 28.21 &  9.38 &  1.16 &  \textbf{1.09} & 1.29 \\
mo-C & \textbf{3.53} & 24.87 & 25.10 &  4.66 &  7.44 & 5.93 \\
\hline
\end{tabular}
\end{table}
    \vspace{-0.3cm}
\end{center}

Overall, our results show that it is beneficial to apply transfer learning to a pretrained VisionNeRF model to obtain a model that explicitly maps grasp poses to grasp success. Furthermore, the results suggest that if a VisionNeRF was trained on multiple objects instead of one, thus learning a more descriptive representation of the scene, the obtained grasp success estimator is also better. While single-grasp models are not able to generalize very well to novel objects, the multi-grasp model, which was trained on the multi-object-B object set performed reasonably well on objects from the multi-object-A set, containing partly similar objects, and also on objects from the multi-object-C set containing objects of different shapes and colors. Considering all tasks, the best model is the combination of the multi-NeRF and the multi-grasp model achieving an average error of 3mm.

\section{Limitations and Future Work}
The VisionNeRFs we train only uses three camera perspectives, leading to distorted rendering for other perspectives. This shows, that the learned scene representation is far from perfect. Our method would most likely benefit from a NeRF that is trained with many perspectives. This however, also leads to a major limitation, as such a model would take a huge effort to obtain in a real world scenario. A possible future work is investigating the possibilities of creating such a model applying sim-2-real transfer learning methods, thus reducing the real world data required.

An other strong limitation is, that our experiments only consider 3DoF grasps and only in simulation. Exploring 5DoF and 6DoF grasps, especially in a real-world experiment, is crucial for a successful model architecture, and is thus also in the scope of possible future works.

For training the grasp estimation models, we used 100 demonstrations from each task. On one hand this does not seem to be that many, if we consider that deep learning architectures usually require an exceptionally large body of data to train on, however for a real world application it would be beneficial if one could reduce the amount of demonstrations to the minimum while retaining the robustness of the method.

\section{Conclusion}
In this work, we propose a unique approach to robotic grasping, employing transfer learning on a trained VisionNeRF to explicitly map grasp poses to their corresponding grasp success. We further applied a gradient-based optimization method on this learned mapping to refine the poses of grasp candidates and thereby attain successful grasp poses. We demonstrated the efficacy of our method on four simulated 3DoF robotic grasping tasks, and showed its ability to generalize to novel objects.

A clear direction for future work is the extension of our method to 5DoF and 6DoF grasps, and to apply it on real world tasks. The methodology we propose here is not limited to robotic grasping, but can be extended to estimate the success of other types of robotic manipulation or interaction. An intriguing prospect for future development of our work could involve integrating additional criteria, tailored to specific tasks, into the optimization objective. One such criterion could be language conditioning, this could provide a foundation for robots to handle tasks of greater complexity.

\bibliographystyle{spmpsci} 
\bibliography{llncs}

\end{document}